\documentclass[conference]{IEEEtran}
\usepackage{graphicx}
\usepackage{amsmath}
\usepackage{amssymb}
\usepackage{tabularx}
\usepackage{cite}
\usepackage{array}
\usepackage{caption}
\usepackage{amsmath,amssymb,amsfonts}
\usepackage{algorithmic}
\usepackage{textcomp}
\usepackage{xcolor}
\ifCLASSINFOpdf
\else
\fi
\begin{document}
%
\title{Integrating Features for Recognizing Human Activities through Optimized Parameters in Graph Convolutional Networks and Transformer Architectures}

\author{\IEEEauthorblockN{Mohammad Belal}
\IEEEauthorblockA{Department of Mechanical and\\Nuclear Engineering\\
Khalifa University\\
Abu Dhabi, United Arab Emirates\\
Email: 100062548@ku.ac.ae}
\and
\IEEEauthorblockN{Taimur Hassan}
\IEEEauthorblockA{Department of Electrical and\\Computer Engineering\\
Abu Dhabi University\\
Abu Dhabi, United Arab Emirates\\
Email: taimur.hassan@adu.ac.ae}
\and
\IEEEauthorblockN{Abdelfatah Hassan}
\IEEEauthorblockA{Department of Electrical and\\Computer Engineering\\
Khalifa University\\
Abu Dhabi, United Arab Emirates\\
Email: 100059689@ku.ac.ae}
\and
\IEEEauthorblockN{Nael Alsheikh}
\IEEEauthorblockA{Department of Mechanical and\\Nuclear Engineering\\
Khalifa University\\
Abu Dhabi, United Arab Emirates\\
Email: 100062606@ku.ac.ae}
\and
\IEEEauthorblockN{Noureldin Elhendawi}
\IEEEauthorblockA{Department of Mechanical and\\Nuclear Engineering\\
Khalifa University\\
Abu Dhabi, United Arab Emirates\\
Email: noureldin.elhendawi@ku.ac.ae}
\and
\IEEEauthorblockN{Irfan Hussain}
\IEEEauthorblockA{Department of Mechanical and\\Nuclear Engineering\\
Khalifa University\\
Abu Dhabi, United Arab Emirates\\
Email: irfan.hussain@ku.ac.ae}}


%


\maketitle

\begin{abstract}
Human activity recognition is a major field of study that employs computer vision, machine vision, and deep learning techniques to categorize human actions. The field of deep learning has made significant progress, with architectures that are extremely effective at capturing human dynamics. This study emphasizes the influence of feature fusion on the accuracy of activity recognition. This technique addresses the limitation of conventional models, which face difficulties in identifying activities because of their limited capacity to understand spatial and temporal features. The technique employs sensory data obtained from four publicly available datasets: HuGaDB, PKU-MMD, LARa, and TUG. The accuracy and F1-score of two deep learning models, specifically a Transformer model and a Parameter-Optimized Graph Convolutional Network (PO-GCN), were evaluated using these datasets. The feature fusion technique integrated the final layer features from both models and inputted them into a classifier. Empirical evidence demonstrates that PO-GCN outperforms standard models in activity recognition. HuGaDB demonstrated a 2.3\% improvement in accuracy and a 2.2\% increase in F1-score. TUG showed a 5\% increase in accuracy and a 0.5\% rise in F1-score. On the other hand, LARa and PKU-MMD achieved lower accuracies of 64\% and 69\% respectively. This indicates that the integration of features enhanced the performance of both the Transformer model and PO-GCN.
\end{abstract}
%


%
\IEEEpeerreviewmaketitle

\section{Introduction}
Human activity recognition is an important component in the science of computer vision, as it is responsible for identifying and categorizing human actions collected on video. The goal of this field is to decode and learn human interactions in video sequences for a variety of applications ranging from surveillance systems to proficiency evaluations \cite{Memon2021-wa}. Particularly in the context of assistive robotic exoskeletons, this topic is useful in a variety of applications including video material categorization and enabling instantaneous and accurate changes in reaction to human motions \cite{Kumar2021-hm}\cite{Fang2020-et}. Another study focused on several strategies for human activity recognition, with deep learning-based approaches significantly improving the precision and efficacy of these recognition systems \cite{De_Miguel-Fernandez2023-rh} \cite{Rahayu2023-ya}.

Deep learning methods combined with sensor-derived data have produced complex models able to evaluate video-based human activities. These models have shown promise in enhancing the interface between human movement and technologies like exoskeletons, enabling exact and timely changes in reaction to human motions \cite{10556541}. Minimizing latency in the connection between human motions and the associated adjustments in mechanical aids while maintaining accuracy and integrity, is an important area of research. Deep learning approaches have proved essential in resolving the challenges inherent in this element of human-robot interaction \cite{Singhania2022-vn, Vrigkas2015-zx, belal2024feature}.

\section{Related Work}
\label{sec:Rel}

A lot of study has lately been done on how to identify individuals based on their movements. Using computer algorithms, several researchers—including Mohsen—have developed fresh approaches for estimating these activities, and the results show great promise \cite{Mohsen2023-dx}. Others used ready-made systems including ResNet50 and ViT, and they were practically always able to accurately spot activity  \cite{Surek2023-cw}. Huang et al. \cite{9500219} introduced a new method offering better knowledge of activities by concentrating on the movement sequence over time. Research on using several approaches in deep learning to learn from less data while nonetheless precisely identifying actions has also been done. Learning the spatial and temporal elements of the human body's movement. In \cite{9998567} introduced a technique for separating activities into smaller pieces. This lets computers decide what individuals are doing more precisely. Moreover, in \cite{Liu2023-mk} a new approach to identify actions via feature diffusion is suggested, which uses a different approach to predict activities depending on video data. This approach works well for addressing problems of recognizing activities over time, especially in terms of knowledge of where the activity starts and finishes as well as the relationships between multiple acts.

Deep learning models still have difficulties even if they have pushed limits in identifying human activity, they cannot completely grasp the space and timing of movements. This work makes important strides in overcoming these issues:

\begin{itemize} 
\item \textbf{Diverse data sources}: We train and test two types of models: the PO-GCN and a Transformer using information from four distinct datasets.
\item \textbf{Comparing models}:Examining how the Transformer and PO-GCN perform on every dataset helps us to identify areas of strength and areas where each model may be improved. 
\item \textbf{Merging features}: Combining several types of data can help identify what activity someone is engaged in, therefore facilitating more accurate and dependable systems, according to our researcher. 
\item \textbf{Taking the advantages of two deep learning models}: We take advantage of the Transformer's ability to understand temporal patterns and the PO-GCN's ability in obtaining complex spatial and temporal features. This shows that the use of multiple models combined together produces better results. 
\end{itemize}

\section{Proposed Method}
\label{sec:pagestyle}
This section describes our proposed approach. We employed four different datasets—HuGaDB, PKU-MMD, LARa, and TUG—to develop two different models: the PO-GCN and a Transformer, with the purpose of detecting human activity. We retrieved features from the last layers of both models and combined them using concatenation. These merged features were then fed into a Fully Connected Network classifier for the final classification.
\begin{figure*}[ht]
    \centering
    \includegraphics[width=1\textwidth]{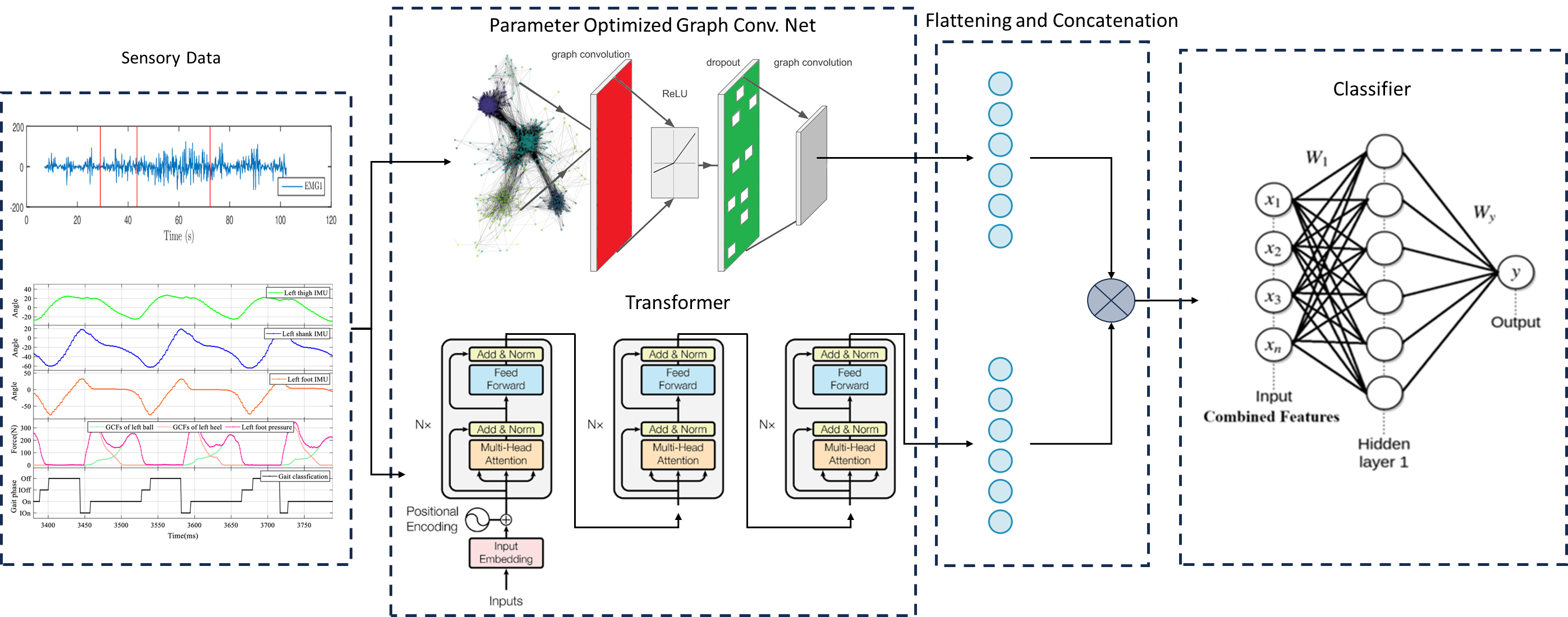}
    \caption{The features were extracted from the GCN and Transformer, then two processes were used which are flattened and concatenated. The classifier received the combined features as inputs.}
    \label{figure_1}
\end{figure*}
\subsection{Parameter-Optimized Graph Convolutional Network}
This part describes how the PO-GCN model, meant to find activities based on human skeletal structure, has shown good performance in several studies. Because the PO-GCN has been refined through parameter changes, its architecture is outstanding in capturing the temporal and spatial correlations present in human movement. Layers in every numerous stage of the model process and transmit data across the spatiotemporal graph of human skeletal movements using graph convolutional networks (GCN). It generates spatial and temporal convolutions at every stage. After that, the revised feature mappings move to the next stage; thereafter, a graph pooling phase reduces the size of the graph.

Cross-entropy loss (CE) and mean squared error (MSE) loss were the two loss functions we used in the training phase. Defined as the CE loss—summed over all phases—this is:
\begin{equation}
\begin{aligned}
& C E=\sum_{stage=1}^{stage} C E_{stages, class}
\end{aligned}
\end{equation}
\\ \begin{equation}
\begin{aligned}
& C E_{class}=\frac{-1}{N} \sum_n y_{n, c} \log \left(\hat{y}_{n, c}\right),
\end{aligned}
\end{equation}
where CE shows the total loss over all stages where $CE_{\text{class}}$ reflects the loss between the actual label $y_{\text{n,c}}$ and the predicted likelihood $\hat{y}_{\text{n,c}}$ for class (c) at sample (n). Usually applied for regression problems, the MSE loss comes from:
\begin{equation}
M S E=\frac{1}{N} \sum_{i=1}^N\left(y_i-\hat{y}_i\right)^2
\end{equation}
where $y_{\text{i}}$ is the true value; N is the sample count; and $\hat{y}_{\text{i}}$ is the expected value for sample (i). These two losses taken together produce the loss function:
\begin{equation}
L_{\text {Total }}=\sum_{stage=1}^{stage} C E+\sigma M S E \text {, }
\end{equation}
where $L_{\text {Total }}$ is the overall combined loss; sigma ($\sigma$) is the weight given to the MSE loss inside this combined function. Combining loss functions aims mostly to produce predictions with less over-recognition mistakes. The model is flexible enough for many applications of activity recognition since its last output assigns action labels to every frame in the sequence.
\subsection{Transformer}
We reported in this work a Transformer-based model for human activity recognition. The identical set of hyperparameters applied for the PO-GCN was applied to train this model. Making use of the Transformer design has several advantages. It is mostly better in extracting features since it can handle data inputs in their whole instead of in segments. The model is able to find complex patterns that it might miss if it only looked at individual parts because it looks at the whole picture.

Moreover, transformers are quite famous for their capacity to precisely depict long-range dependencies in data. This shows how well the model understands the link between activities occurring at different times, which is crucial for correctly analyzing movement sequences concerning human activity detection. By considering these temporal connections, the Transformer model may more precisely project future actions or the continuation of an activity \cite{Kumar2024-sl}.
\subsection{Last Layer Feature Fusion}
Feature fusion, sometimes referred to as late fusion, is a fundamental method in deep learning whereby features from several models are combined to improve the general performance of the model \cite{6918213}. Using two separate models—the Transformer and the Parameter-Optimized GCN—activity recognition helped to extract features from their last layers \cite{Zhang2021-io} \cite{Mungoli2023-hl}. The next method applied to mix the obtained features was concatenation. This approach can perhaps increase the general representational and predictive strength by combining the several insights acquired by every particular model. Fundamental ideas in deep learning and machine learning, feature fusion has been extensively investigated in academic publications, where two separate models were used in the framework of this work for feature extraction from their last layers: Transformer and Parameter-Optimized GCN. Concatenation then helped to combine the retrieved features. This method helps to combine different and complementary insights gained by every particular model, hence perhaps increasing the general representational and predictive power. With much research in academic publications, feature fusion is a pillar in both the machine and deep learning fields. Researchers have proposed several fusion strategies including attentional feature fusion and guided training in an attempt to enhance classification task performance \cite{Dai_undated-ji}.
\subsection{Classifier}
This research uses the fully connected network meant to analyze the fused feature set obtained from the combined outputs of the Transformer and Parameter-Optimized GCN models as a classifier \cite{Basha2019-ji} \cite{Kalayci2022-cs}. The architecture of this network consists of several layers cooperating to maximize classification performance. Its batch normalizing layer helps every neuron to standardize the inputs, so reducing internal covariate shift and improving neural network stability. Two dense layers follow from this: a final output layer producing the classification results and a flattening layer transforming the multi-dimensional input data features into a one-dimensional array so enabling the dense layers to handle it.

Deep learning algorithm Adaptive Moment Estimation (Adam) is the highly esteemed optimizer used in the network \cite{Kingma2014-at}. Adam is well-known in real-time parameter changes that significantly raise the model's learning rate and accuracy. It comprises an adjustable learning rate that changes with learning progress and a momentum component pushing the optimizer in the proper direction.
Consequently, the convergence toward the optimal parameter selection that reduces the loss function can happen faster.

Most often used loss function in classification systems is cross-entropy (CE) loss. By means of a comparison between the predicted and actual probability distribution, this loss function measures the performance of the model, therefore rendering it suitable for classification problems. The degree to which the expected likelihood differs from the actual label determines how much the CE loss rises. By reducing the CE loss during training, the model teaches it to generate predictions that closely match the true labels and so improves classification accuracy.

From feature fusion until the final classification output, figure \ref{figure_1} offers a clear graphic of the system architecture. This graph offers direction on how the several components of the network interact to produce a logical and efficient classifier.
\section{Network Setup}
\label{sec:typestyle}
\subsection{Public Datasets}
\subsubsection{Human Gait Database (HuGaDB)}
Human gait data is gathered in the HuGaDB collection and applied for activity recognition \cite{Chereshnev2018-nb}. Among the several activities included in the collection are continuous recordings of sitting, running, walking, climbing and descending stairs. Apart from two electromyography sensors tracking muscle activation, wearable inertial sensors on the thighs, shins, and feet were employed to gather data. This dataset simultaneously serves two purposes: it provides information on the relative movements of various leg components and aids in the identification of various activities by means of which they are executed.
\subsubsection{PKU-MMD}
a vast action recognition resource with 1,076 lengthy video sequences in 51 action categories. 66 different subjects participated in three separate camera positions to document these actions \cite{Liu2017-pn}. Comprising close to 20,000 action instances and 5.4 million frames, the PKU-MMD dataset Captured using the Kinect v2 sensor it offers multi-modal data streams including RGB, depth, infrared radiation, and skeletal data \cite{Liu2017-pn}.
\subsubsection{Logistic Activity Recognition Challenge (LARa)}
First openly available dataset targeted on human activity recognition in logistics environments: LARa, logistic activity recognition challenge \cite{Niemann2020-rf}. It developed under the Innovation Lab Hybrid Services in Logistics at TU Dortmund University. Recorded using motion capture, inertial measurement units, and RGB cameras, the dataset comprises recordings of 14 individuals engaged in selecting and packing jobs. The dataset consists in 758 minutes of well labeled records spanning 474 person-hours by 12 annotators. The data consists of eight activity classifications and nineteen binary semantic features \cite{Niemann2020-rf}.
\subsubsection{Timed Up and Go}
A well-known assessment tool for measuring the mobility, balance, walking capacity, and fall risk of an older adult \cite{Podsiadlo1991-tc}. The patient gets out of a chair, moves three meters at a comfortable and safe pace, turns around, moves back to the chair, and then settles in to finish the test. Time it takes a patient to complete this activity will help doctors ascertain their functional mobility and fall risk. In the realm of geriatric care, the TUG test is a priceless tool since it provides vital information about the physical abilities and fall risk of an elderly person \cite{Podsiadlo1991-tc}.
\subsection{Metrics}
Using a range of conventional metrics, including F1-score and general accuracy, the suggested model's performance was evaluated. Particular application of the F1-score for segment-wise assessment using \cite{Farha2019-ke} where every anticipated action segment is categorized as either a true positive (TP) or a false positive (FP) \cite{Lea2017-re}.The F1 score computed follows:
\begin{equation}
F1-\text {score}=\frac{T P}{T P+0.5(F N+F P)}
\end{equation}
One classification evaluation measure is accuracy \cite{Farha2019-ke}. Since accuracy is measured by the number of properly classified cases, it can be used to test classification tasks when the classes are balanced, which means that each has almost the same number of samples. It offers a direct evaluation of the efficacy of a model and is easily comprehensible. Accuracy is defined as:
\begin{equation}
Acc =\frac{T P+T N}{T P+T N+F P+F N}
\end{equation}
These metrics are for every action category, these measures were computed segmentally.

\section{Results \& Discussion}
\label{sec:majhead}
This work aims to evaluate and compare the performance of Transformer model and other top activity identification models with Parameter-Optimized GCN. Evaluating the robustness and efficacy of our proposed method was the aim. This study aims to improve the awareness of the advantages and drawbacks inherent in every model by way of a comparison examination. Table \ref{table1} shows the comparative findings—that is, the results of running the models for 100 epochs with a batch size of 4 considering the variations in sampling rates among the four datasets. Included among the outcomes were F1 scores and accuracy.
\begin{table}[h]
\raggedleft
\caption{The F1-score and Accuracy values for the PO-GCN and Transformer activity recognition models.}
\captionsetup{skip=1ex}
\begin{tabularx}{1\linewidth}{|>{\centering\arraybackslash}p{1.4cm}|>{\centering\arraybackslash}p{0.9cm}|>{\centering\arraybackslash}p{1.59cm}|>{\centering\arraybackslash}p{1.15cm}|>{\centering\arraybackslash}p{1.62cm}|}
\hline \textbf{Architecture} & \multicolumn{2}{|c|}{ \textbf{PO-GCN} } & \multicolumn{2}{c|}{ \textbf{Transformer} } \\
\hline \textbf{Dataset} & \textbf{Acc\%} & \textbf{F1\%} & \textbf{Acc\%} & \textbf{F1\%}\\
\hline HuGaDB & 92.7 & 95.2 & 90.3 & 94.0 \\
 LARa & 64.3 & 40.6 & 59.3 & 30.5 \\
 PKU-MMD & 69.0 & 48.2 & 68.3 & 52.9 \\
 TUG & 93.2 & 98.3 & 90.9 & 98.1 \\ \hline

\end{tabularx}
\label{table1}
\end{table}

\begin{table}[h]
\centering
\caption{Comparison of findings from \cite{9998567}, PO-GCN, and other activity recognition models}
\captionsetup{skip=1ex}
\begin{tabularx}{0.996\linewidth}
{|>{\centering\arraybackslash}p{2.49cm}|>{\centering\arraybackslash}p{2.49cm}|>{\centering\arraybackslash}p{2.51cm}|}
\hline 
\textbf{HuGaDB} & \textbf{Acc\%} & \textbf{F1-score\%}\\ 
\hline
MS-GCN & 90.4 & 93.0\\
ST-GCN & 88.7 & 67.7\\
Transformer & 90.3 & 94.0\\
PO-GCN & \textbf{92.7} & \textbf{95.2}\\
\hline
\textbf{LARa} & \textbf{Acc\%} & \textbf{F1-score\%}\\ 
\hline
MS-GCN & 65.6 & \textbf{43.6}\\
ST-GCN & \textbf{67.9} & 25.8\\
Transformer & 59.3 & 30.5\\
PO-GCN & 64.31 & 40.63\\
\hline
\textbf{PKU-MMD} & \textbf{Acc\%} & \textbf{F1-score\%}\\
\hline
MS-GCN & 68.5 & 51.6\\
ST-GCN & 64.9 & 15.5\\
Transformer & 68.3 & \textbf{52.9}\\
PO-GCN & \textbf{69.0} & 48.16\\
\hline
\textbf{TUG} & \textbf{Acc\%} & \textbf{F1-score\%}\\ 
\hline
MS-GCN & \textbf{93.6} & 97.9\\
ST-GCN & 93.2 & 93.8\\
Transformer & 90.9 & \textbf{98.1}\\
PO-GCN & 93.2 & \textbf{98.3}\\
\hline
\end{tabularx}
\label{table2}
\end{table}

\begin{table}[h]
\raggedleft
\caption{Results from feature fusion using a combination of features from the PO-GCN and the Transformer.}
\captionsetup{skip=1ex}
\begin{tabularx}{1\linewidth}
{|m{2.21cm}|m{2.71cm}|m{2.61cm}|m{2.6cm}|m{2.6cm}|}
\hline \textbf{Method} & \multicolumn{2}{|c|}{ \textbf{Last layer Fusion}} \\
\hline \textbf{Dataset} & \textbf{Acc\%} & \textbf{F1\%}\\
\hline HuGaDB  & 84.70 & 88.20 \\
 LARa  & 59.30 & 50.48 \\
 PKU-MMD  & 96.61 & 94.95 \\
 TUG  & 98.44 & 97.66 \\ \hline

\end{tabularx}
\label{table_3}
\end{table}

A comparison of the Transformer model, the suggested PO-GCN model, the performance results reported by \cite{9998567}, and other activity recognition techniques across multiple datasets is shown in Table \ref{table2}. The PO-GCN obtained an F1-score of 95.2\% and an impressive accuracy of 92.7\% for the HuGaDB dataset. In comparison, the PO-GCN's F1-score of 40.63\% was 3\% lower than the state-of-the-art on the LARa dataset, and its accuracy of 64.31\% was 3.6\% lower than the ST-GCN model. Fascinatingly, the PO-GCN demonstrated a minor increase in accuracy on the PKU-MMD dataset, hitting 69\%; however, its F1-score of 48.16\% was 3.4\% less than the Transformer model, which obtained the highest F1-score of 52.9\%. With an F1-score of 98.3\% for the TUG dataset, the PO-GCN performed well, albeit its accuracy of 93.2\% behind the state-of-the-art. On the PKU-MMD dataset, the Transformer model performed fairly well, coming in close to the reported values for HuGaDB, but it was much less accurate than other models and PO-GCN. The Transformer had a lower accuracy on the TUG dataset, but the same F1-score as the best models.

The results of the feature fusion method for human activity recognition are presented in Table \ref{table_3}. Feature fusion was used to increase accuracy and F1-score while displaying adaptability. Feature fusion produced an F1-score of 50.48\% in the LARa dataset. The PKU-MMD dataset showed that feature fusion performed better than PO- GCN alone, yielding an accuracy of 96.61\% and an F1- score of 94.95\%. Additionally, the feature fusion approach increased accuracy for the TUG dataset, achieving 98.4\%. According to these findings, the feature fusion approach may be a useful tool for raising activity recognition model performance, especially when it comes to increasing accuracy and F1-score across various datasets.

 \section{Conclusion}
\label{sec:conc}
This work presents a novel approach for recognizing human actions by means of data from four separate sources. We thus applied two models: PO-GCN and Transformer model. Following considerable training and evaluation on the datasets, each model's efficacy was assessed using accuracy and F1-score measures. We used the unique features of transformers and graph convolutional networks by means of a deep learning method termed feature fusion. This approach aggregates last layer characteristics from both models before feeding them into a classifier, hence enhancing performance.
Regarding activity recognition, the PO-GCN beats other models according to the data. Furthermore, in three of the datasets the feature fusion method exceeded the stand-alone PO-GCN based on recognition rates. Together in this approach, the Transformer's capacity to process long-term patterns and the sensitivity of the PO-GCN to temporal and spatial details complement each other.



%


\begin{thebibliography}{9}
\bibitem{Memon2021-wa}
F.~A. Memon, U.~A. Khan, A.~Shaikh, A.~Alghamdi, P.~Kumar, and M.~Alrizq,
  ``Predicting actions in videos and action-based segmentation using deep
  learning,'' \emph{IEEE Access}, vol.~9, pp. 106918--106932, 2021.

\bibitem{Fang2020-et}
B.~Fang, Q.~Zhou, F.~Sun, J.~Shan, M.~Wang, C.~Xiang, and Q.~Zhang, ``Gait
  neural network for human-exoskeleton interaction,'' \emph{Front. Neurorobot.},
  vol.~14, p.~58, Oct. 2020.
\bibitem{De_Miguel-Fernandez2023-rh}
J.~de Miguel-Fern{\'a}ndez, J.~Lobo-Prat, E.~Prinsen, J.~M. Font-Llagunes, and
  L.~Marchal-Crespo, ``Control strategies used in lower limb exoskeletons for
  gait rehabilitation after brain injury: a systematic review and analysis of
  clinical effectiveness,'' \emph{J. Neuroeng. Rehabil.}, vol.~20, no.~1, p.~23,
  Feb. 2023.
\bibitem{9998567}
B.~Filtjens, B.~Vanrumste, and P.~Slaets, ``Skeleton-based action segmentation
  with multi-stage spatial-temporal graph convolutional neural networks,''
  \emph{IEEE Transactions on Emerging Topics in Computing}, pp. 1--11, 2022.

\bibitem{Liu2023-mk}
D.~Liu, Q.~Li, A.~Dinh, T.~Jiang, M.~Shah, and C.~Xu, ``Diffusion action
  segmentation,'' \emph{arXiv}, 2023.
\bibitem{9500219}
F.~A. Memon, U.~A. Khan, A.~Shaikh, A.~Alghamdi, P.~Kumar, and M.~Alrizq,
  ``Predicting actions in videos and action-based segmentation using deep
  learning,'' \emph{IEEE Access}, vol.~9, pp. 106918--106932, 2021.

\bibitem{Singhania2022-vn}
D.~Singhania, R.~Rahaman, and A.~Yao, ``{C2F-TCN}: A framework for semi and
  fully supervised temporal action segmentation,'' \emph{arXiv}, Dec. 2022.
\bibitem{Kumar2021-hm}
S.~Kumar, S.~Haresh, A.~Ahmed, A.~Konin, M.~Z. Zia, and Q.-H. Tran,
  ``Unsupervised action segmentation by joint representation learning and
  online clustering,'' \emph{arXiv}, 2021.

\bibitem{Chereshnev2018-nb}
R.~Chereshnev and A.~Kert{\'e}sz-Farkas, ``{HuGaDB}: Human gait database for
  activity recognition from wearable inertial sensor networks,'' in \emph{Lecture
  Notes in Computer Science}, Cham: Springer International Publishing, 2018, pp.
  131--141.
\bibitem{Niemann2020-rf}
F.~Niemann, C.~Reining, F.~Moya Rueda, N.~R. Nair, J.~A. Steffens, G.~A. Fink,
  and M.~Ten Hompel, ``{LARa}: Creating a dataset for human activity recognition
  in logistics using semantic attributes,'' \emph{Sensors (Basel)}, vol.~20,
  no.~15, p. 4083, Jul. 2020.

\bibitem{Liu2017-pn}
C.~Liu, Y.~Hu, Y.~Li, S.~Song, and J.~Liu, ``{PKU-MMD}: A large scale benchmark
  for continuous multi-modal human action understanding,'' \emph{arXiv}, Mar.
  2017.
\bibitem{Podsiadlo1991-tc}
D.~Podsiadlo and S.~Richardson, ``The timed ``Up \& go'': A test of basic
  functional mobility for frail elderly persons,'' \emph{J. Am. Geriatr. Soc.},
  vol.~39, no.~2, pp. 142--148, Feb. 1991.

\bibitem{Farha2019-ke}
Y.~A. Farha and J.~Gall, ``{MS-TCN}: {Multi-Stage} Temporal Convolutional Network
  for Action Segmentation,'' in \emph{2019 {IEEE/CVF} Conference on Computer Vision
  and Pattern Recognition ({CVPR})}, Long Beach, CA, USA, Jun. 2019.

\bibitem{Lea2017-re}
C.~Lea, M.~D. Flynn, R.~Vidal, A.~Reiter, and G.~D. Hager, ``Temporal
  Convolutional Networks for Action Segmentation and Detection,'' in \emph{2017
  {IEEE} Conference on Computer Vision and Pattern Recognition ({CVPR})},
  Honolulu, HI, Jul. 2017.
\bibitem{6918213}
S.~Khalid, T.~Khalil, and S.~Nasreen, ``A survey of feature selection and
  feature extraction techniques in machine learning,'' in \emph{2014 Science and
  Information Conference}, 2014, pp. 372--378.

\bibitem{Zhang2021-io}
T.~Zhang, S.~Fan, J.~Hu, X.~Guo, Q.~Li, Y.~Zhang, and A.~Wulamu, ``A feature
  fusion method with guided training for classification tasks,'' \emph{Comput.
  Intell. Neurosci.}, vol. 2021, p. 6647220, Apr. 2021.

\bibitem{Mungoli2023-hl}
N.~Mungoli, ``Adaptive Feature Fusion: Enhancing generalization in deep learning
  models,'' \emph{arXiv}, 2023.
  
\bibitem{Dai_undated-ji}
Y.~Dai, F.~Gieseke, S.~Oehmcke, Y.~Wu, and K.~Barnard, ``Attentional Feature
  Fusion,'' accessed: Nov. 21, 2023.

\bibitem{Basha2019-ji}
S.~H.~S. Basha, S.~R. Dubey, V.~Pulabaigari, and S.~Mukherjee, ``Impact of Fully
  Connected layers on performance of Convolutional Neural Networks for image
  classification,'' \emph{arXiv}, 2019.
\bibitem{Kalayci2022-cs}
T.~A. Kalayc{\i} and U.~Asan, ``Improving classification performance of fully
  connected layers by Fuzzy clustering in transformed feature space,''
  \emph{Symmetry (Basel)}, vol.~14, no.~4, p. 658, Mar. 2022.

\bibitem{Kingma2014-at}
D.~P. Kingma and J.~Ba, ``Adam: A method for stochastic optimization,''
  \emph{arXiv}, 2014.
  
\bibitem{Rahayu2023-ya}
E.~S. Rahayu, E.~M. Yuniarno, I.~K. E. Purnama, and M.~H. Purnomo, ``Human
  activity classification using deep learning based on {3D} motion feature,''
  \emph{Mach. Learn. Appl.}, vol.~12, no. 100461, p. 100461, Jun. 2023.

\bibitem{Vrigkas2015-zx}
M.~Vrigkas, C.~Nikou, and I.~A. Kakadiaris, ``A review of human activity
  recognition methods,'' \emph{Front. Robot. AI}, vol.~2, Nov. 2015.
\bibitem{Mohsen2023-dx}
S.~Mohsen, ``Recognition of human activity using {GRU} deep learning algorithm,''
  \emph{Multimed. Tools Appl.}, vol.~82, no.~30, pp. 47733--47749, Dec. 2023.

\bibitem{10118525}
R.~Moola and A.~Hossain, ``Human Activity Recognition using Deep Learning,'' in
  \emph{2022 URSI Regional Conference on Radio Science (USRI-RCRS)}, 2022, pp.
  1--4.
\bibitem{Surek2023-cw}
G.~A.~S. Surek, L.~O. Seman, S.~F. Stefenon, V.~C. Mariani, and L.~D.~S. Coelho,
  ``Video-based human activity recognition using deep learning approaches,''
  \emph{Sensors (Basel)}, vol.~23, no.~14, Jul. 2023.

\bibitem{Kumar2024-sl}
P.~Kumar, S.~Chauhan, and L.~K. Awasthi, ``Human activity recognition ({HAR})
  using deep learning: Review, methodologies, progress and future research
  directions,'' \emph{Arch. Comput. Methods Eng.}, vol.~31, no.~1, pp. 179--219,
  Jan. 2024.
\bibitem{belal2024feature}
M.~Belal, T.~Hassan, A.~Ahmed, A.~Aljarah, N.~Alsheikh, and I.~Hussain, ``Feature
  Fusion for Human Activity Recognition using Parameter-Optimized Multi-Stage
  Graph Convolutional Network and Transformer Models,'' \emph{arXiv}, 2024.

\bibitem{10556541}
M.~Belal, N.~Alsheikh, A.~Aljarah, and I.~Hussain, ``Deep Learning Approaches for
  Enhanced Lower-Limb Exoskeleton Control: A Review,'' \emph{IEEE Access}, pp.
  1--1, 2024.
\end{thebibliography}

\end{document}